\documentclass[conference]{IEEEtran}
\usepackage{cite}
\usepackage{amsmath,amssymb,amsfonts}
\usepackage{algorithmic}
\usepackage{graphicx}
\usepackage{textcomp}
\usepackage{xcolor}
\usepackage{hhline}
\usepackage{color, colortbl}
\usepackage{multirow}
\usepackage{times}
\usepackage{epsfig}
\usepackage{graphicx}
\usepackage{amsmath}
\usepackage{amssymb}
\usepackage{subcaption}
\usepackage[utf8]{inputenc}
\usepackage{lipsum} 
\usepackage{booktabs}
\usepackage{xcolor}
\usepackage{makecell}
\usepackage{url}
\usepackage[bookmarks=false]{hyperref}

\def\BibTeX{{\rm B\kern-.05em{\sc i\kern-.025em b}\kern-.08em
    T\kern-.1667em\lower.7ex\hbox{E}\kern-.125emX}}
\begin{document}


\title{A Method for Curation of Web-Scraped \\Face Image Datasets}

\author{\IEEEauthorblockN{Kai Zhang, Vítor Albiero and Kevin W. Bowyer}
\IEEEauthorblockA{\textit{Department of Computer Science \& Engineering, University of Notre Dame}}
{\tt\small \{kzhang4, valbiero, kwb\}@nd.edu}}

\maketitle

\begin{abstract}
Web-scraped, in-the-wild datasets have become the norm in face recognition research.
The numbers of subjects and images acquired in web-scraped datasets are usually very large, with number of images on the millions scale.
A variety of issues occur when collecting a dataset in-the-wild, including images with the wrong identity label, duplicate images, duplicate subjects and variation in quality.
With the number of images being in the millions, a manual cleaning procedure is not feasible. 
But fully automated methods used to date result in a less-than-ideal level of clean dataset.
We propose a semi-automated method, where the goal is to have a clean dataset for testing face recognition methods,
with similar quality across men and women, to support comparison of accuracy across gender.
Our approach removes near-duplicate images, merges duplicate subjects, corrects mislabeled images, and removes images outside a defined range of pose and quality.
We conduct the curation on the Asian Face Dataset (AFD) and VGGFace2 test dataset.
The experiments show that a state-of-the-art method achieves a much higher accuracy on the datasets after they are curated.
Finally, we release our cleaned versions of both datasets to the research community.
\end{abstract}


\section{Introduction}

The advance of deep-learning-based face recognition methods is largely due to the availability of large face datasets, which are generally created by scraping the web for images.
Mislabeled images, flipped identity labels, duplicate images, and duplicate subjects are common problems in web-scraped datasets.
The larger the web-scraped dataset is, the lower the chances that it is fully curated.

Wang et al.~\cite{wang2018devil} reported on the impact that web-scraped dataset problems have on training datasets.
The authors showed that models trained on manually cleaned subsets of two popular datasets, containing only about 32\% and 20\% of the original data, achieved comparable accuracy with models that were trained on the entire dataset.
Other works~\cite{wang2019comining, zhong2019unequal} also propose practices to handle the noisy data inequality to achieve better training results.

Cao et al.~\cite{vggface2} released a training and testing dataset focused on pose and age heterogeneity.
The authors performed a semi-automated cleaning that consisted of several steps, but the dataset still contains instances of the aforementioned problems after this curation.

When comparing accuracy of different methods, or especially when comparing accuracy differences across demographics, 
the testing dataset must be as clean as possible, and as similar in quality as possible across the demographics being studied.
This setting was more naturally true of datasets that were constructed from images acquired in controlled scenarios,
but controlled-acquisition datasets are prohibitively expensive to create at the same scale as web-scraped datasets.

The FERET dataset is perhaps the best-known early face image dataset to be widely used in the research community \cite{Phillips_1998, Phillips_2000}.
Images in FERET were collected under controlled conditions and various meta-data was associated with each image at time of acquisition.
FERET was followed by a number of other widely-used datasets that were also collected under controlled conditions (e.g., \cite{frgc}, \cite{MBGC}, \cite{morph}).

In 2007, the Labeled Faces in the Wild (LFW) dataset was released \cite{LFW}.
Departing from the norm up to that time, the images in LFW were not collected under controlled conditions, but rather assembled from already-existing images that represent an ``in-the-wild'' range of pose, illumination, expression and other factors.
LFW contains 13,233 images with a variable number of persons in each image; 1,680 of the persons appear in more than one image and 4,069 appear in a single image.

The popularity of LFW spurred the collection of a succession of larger web-scraped, in-the-wild datasets \cite{ijbc, megaface}, which are mainly composed of Western faces.
To the best of our knowledge, there is not a widely-used, publicly-available dataset with a large number of Asian faces built for face recognition testing.
The best candidate dataset that we are aware of is the Asian Faces Dataset (AFD) \cite{afd}.
The AFD dataset was assembled using in-the-wild images scraped from the web for training purposes. 
Thus it has instances of the same problems that arise in other in-the-wild datasets.

In this paper, we introduce a method to curate datasets to be more appropriate for accuracy testing purposes. 
We repeated the experiment on a second dataset (VGGFace2 testing dataset \cite{vggface2}) and show that on both datasets, the accuracy achieved for both men and women is much higher.
More specifically, we also show that the accuracy before and after curation is higher for men, which agrees with what the literature usually reports \cite{frvt, Lui2009, Beveridge2009, frgc, Grother2010, Klare2012, Grother2017, cook2018, Lu2018, frvt3, albiero2020analysis, albiero2019how, hupont2019demopairs, phillips2009MBGC}.

\begin{figure*}[t]
    \centering
    \begin{subfigure}[b]{.24\textwidth}
        \centering
        \captionsetup[subfigure]{labelformat=empty}
        \begin{subfigure}[b]{.485\columnwidth}
            \centering
            \includegraphics[width=\linewidth]{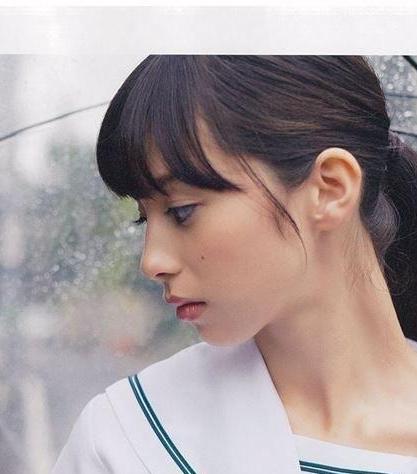}
            \caption{Large pose \\ angle}
        \end{subfigure}
        \begin{subfigure}[b]{.485\columnwidth}
            \centering
            \includegraphics[width=0.9\linewidth]{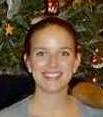}
            \caption{Blurry \\ image}
        \end{subfigure}
        \addtocounter{subfigure}{-2}   
        \vspace{-1em}
        \caption{Low quality images.}
        \vspace{1em}
        \label{fig:analysis_low_quality1}
    \end{subfigure}
    \hfill
    \begin{subfigure}[b]{.24\textwidth}
        \centering
        \captionsetup[subfigure]{labelformat=empty}
        \begin{subfigure}[b]{.485\columnwidth}
            \centering
            \includegraphics[width=\linewidth]{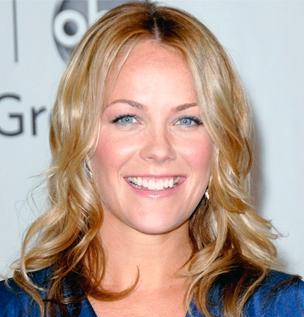}
            \caption{Correct image\\Maud Adams}
        \end{subfigure}
        \begin{subfigure}[b]{.485\columnwidth}
            \centering
            \includegraphics[width=\linewidth]{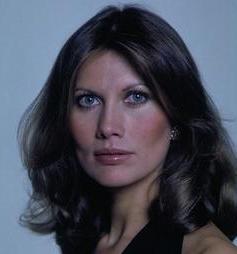}
            \caption{Wrong image\\Andrea Anders}
        \end{subfigure}
        \addtocounter{subfigure}{-2}
        \vspace{-1em}
        \caption{Mislabeled image.}
        \vspace{1em}
        \label{fig:analysis_flipped_id1}
    \end{subfigure}
    \hfill
    \begin{subfigure}[b]{.24\textwidth}
        \centering
        \captionsetup[subfigure]{labelformat=empty}
        \begin{subfigure}[b]{.485\columnwidth}
            \centering
            \includegraphics[width=0.6\linewidth]{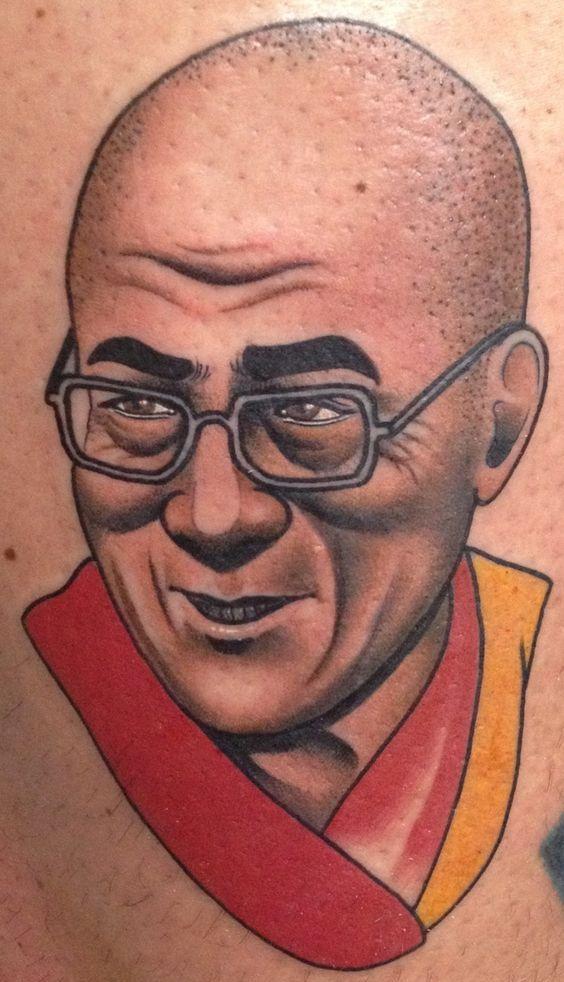}
            \caption{Tattoo \\ Face}
        \end{subfigure}
        \begin{subfigure}[b]{.485\columnwidth}
            \centering
            \includegraphics[width=\linewidth]{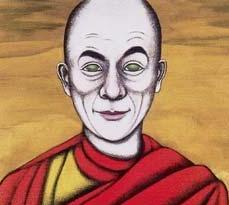}
            \caption{Draw \\ face} 
        \end{subfigure}
        \addtocounter{subfigure}{-2}
        \vspace{-1em}
        \caption{Not real faces.}
        \vspace{1em}
        \label{fig:analysis_duplicated_id1}
    \end{subfigure}
    \hfill
    \vspace{-0.5em}
    \begin{subfigure}[b]{.24\textwidth}
        \centering
        \captionsetup[subfigure]{labelformat=empty}
        \begin{subfigure}[b]{.485\columnwidth}
            \centering
            \includegraphics[width=0.8\linewidth]{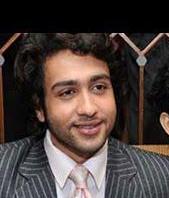}
            \caption{Image \\ 0522}
        \end{subfigure}
        \begin{subfigure}[b]{.485\columnwidth}
            \centering
            \includegraphics[width=\linewidth]{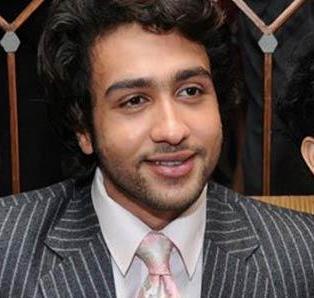}
            \caption{Image \\ 0523} 
        \end{subfigure}
        \addtocounter{subfigure}{-2}
        \vspace{-1em}
        \caption{Near-duplicated images.}
        \vspace{1em}
        \label{fig:analysis_near_duplicate1}
    \end{subfigure}
    \centering
    \begin{subfigure}[b]{.24\textwidth}
        \centering
        \captionsetup[subfigure]{labelformat=empty}
        \begin{subfigure}[b]{.485\columnwidth}
            \centering
            \includegraphics[width=\linewidth]{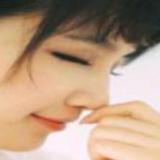}
            \caption{Large pose \\ angle}
        \end{subfigure}
        \begin{subfigure}[b]{.485\columnwidth}
            \centering
            \includegraphics[width=\linewidth]{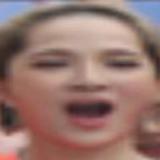}
            \caption{Blurry \\ image}
        \end{subfigure}
        \addtocounter{subfigure}{-2}   
        \vspace{-1em}
        \caption{Low quality images.}
        \label{fig:analysis_low_quality}
    \end{subfigure}
    \hfill
    \begin{subfigure}[b]{.24\textwidth}
        \centering
        \captionsetup[subfigure]{labelformat=empty}
        \begin{subfigure}[b]{.485\columnwidth}
            \centering
            \includegraphics[width=\linewidth]{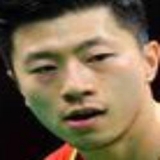}
            \caption{Correct image\\malong}
        \end{subfigure}
        \begin{subfigure}[b]{.485\columnwidth}
            \centering
            \includegraphics[width=\linewidth]{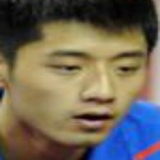}
            \caption{Wrong image\\zhangjike}
        \end{subfigure}
        \addtocounter{subfigure}{-2}
        \vspace{-1em}
        \caption{Flipped identity label.}
        \label{fig:analysis_flipped_id}
    \end{subfigure}
    \hfill
    \begin{subfigure}[b]{.24\textwidth}
        \centering
        \captionsetup[subfigure]{labelformat=empty}
        \begin{subfigure}[b]{.485\columnwidth}
            \centering
            \includegraphics[width=\linewidth]{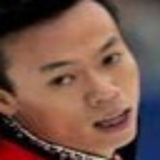}
            \caption{Image identified \\as shenxue}
        \end{subfigure}
        \begin{subfigure}[b]{.485\columnwidth}
            \centering
            \includegraphics[width=\linewidth]{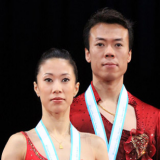}
            \caption{shenxue \\ zhaohongbo} 
        \end{subfigure}
        \addtocounter{subfigure}{-2}
        \vspace{-1em}
        \caption{Duplicated subject identity.}
        \label{fig:analysis_duplicated_id}
    \end{subfigure}
    \hfill
    \begin{subfigure}[b]{.24\textwidth}
        \centering
        \captionsetup[subfigure]{labelformat=empty}
        \begin{subfigure}[b]{.485\columnwidth}
            \centering
            \includegraphics[width=\linewidth]{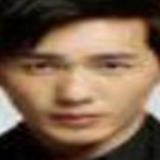}
            \caption{Image \\ 0016}
        \end{subfigure}
        \begin{subfigure}[b]{.485\columnwidth}
            \centering
            \includegraphics[width=\linewidth]{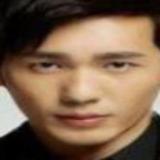}
            \caption{Image \\ 0017} 
        \end{subfigure}
        \addtocounter{subfigure}{-2}
        \vspace{-1em}
        \caption{Near-duplicated images.}
        \label{fig:analysis_near_duplicate}
    \end{subfigure}
    \caption{Common web-scraped issues found in the VGGFace2 test dataset (top) and AFD dataset (bottom).}
    \vspace{-1em}
\end{figure*}

The contributions of this paper can be summarized as follows:
\begin{itemize}
    \item A curated subset of the Asian Faces Dataset (AFD) that is suitable for accuracy comparison tests\footnote{\url{https://github.com/vitoralbiero/afd_dataset_cleaned}};
    \item A new curated version of the VGGFace2 test dataset\footnote{\url{https://github.com/vitoralbiero/vggface2_test_cleaned}};
    \item A comparison of accuracy between men and women in web-scraped datasets;
    \item An approach for how to better curate in-the-wild datasets that can be applied to other datasets.
\end{itemize}

\section{Dataset Analysis}
\label{sec:data_analysis}
The original AFD dataset contains 2,018 subjects and 346,744 images in total.
As the dataset does not contain male and female metadata, we used a gender predictor CNN \cite{insightface} to create gender meta-data. 
For all of the subjects where the gender predictor agreed for 75\% or more of their images, that gender was assigned to the subject.
The rest, around 300 subjects, were manually inspected and a gender assigned.

In initial inspection of the original AFD dataset, we found instances of a variety of problems that we expect occur to some degree in all web-scraped datasets.
There are many images with poor quality on various dimensions, such as images with large face pose angle and large blur, as shown in Figure \ref{fig:analysis_low_quality}. We would like to restrict image qualities such as pose for two reasons. First, If the distribution of images along some quality dimension varies substantially across some demographic, this makes it difficult to reliably compare recognition accuracy across that group. Second, we cannot guarantee different datasets have the same pose distribution without restricting it. For example, the MORPH \cite{morph} dataset has only frontal images.
To more easily compare results on the two datasets, it is useful to restrict the in-the-wild pose to more frontal.

Some images that are assigned to a particular identity 
do not belong to that subject identity.
These images can be from subjects that are not in the dataset (mis-labeled) or from other subjects that are in the dataset ("flipped" identity label).
For example, in AFD, there is a subject named ``malong'', who is a famous table tennis player. 
In the folder of images assigned the ``malong'' label, some images are actually his teammate ``zhangjike'', who is also a subject in the AFD dataset.
Figure \ref{fig:analysis_flipped_id} shows one image from each subject.
The problem of confusion between persons who are ``co-stars'' in a sports team or a movie / television show seems to be a common problem for web-scraped datasets.
Both problems affect testing accuracy, as mislabeled identities will generate supposed authentic matches that have lower similarity scores, and flipped identity labels will, on top of that, generate impostor matches that have higher similarity scores.
Flipped labels were shown to hurt the accuracy of training models significantly more than simple mislabeled images~\cite{wang2018devil}.

Any individual should have only one identity in a dataset, 
but we found that some individuals have multiple assigned identities in AFD.
Some images of certain subjects are wrongly classified as another subject, resulting in the situation where one subject has two different identity names. 
For example, one subject in the AFD dataset is called ``shenxue''.
She is a famous pair mixed skater. 
Her partner, ``zhaohongbo'', is also a subject in AFD. 
However, all the images in the folder of ``shenxue'' are actually of ``zhaohongbo'', as shown in Figure \ref{fig:analysis_duplicated_id}.
Since this couple is always together in photos, we speculate that the main problem was that his face is bigger than hers, which could make the face detector always find his face first.

Lastly, many subjects have multiple near-duplicate images in their folder. 
One example is shown in Figure \ref{fig:analysis_near_duplicate}.
The existence of near-duplicate images will influence the accuracy testing as the authentic matches between the near-duplicate images will have unrealistically high similarity scores.

\begin{figure*}[t]
    \centering
    \includegraphics[width=\linewidth]{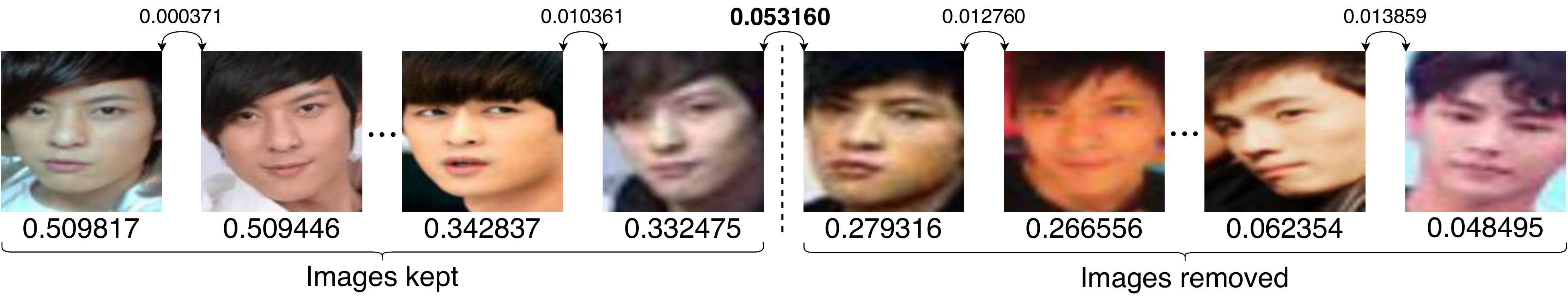}
    \caption{Mislabeled images cleaning overview on one subject. Bottom values represent the mean similarity of the current image to all images of the particular subject, top values represent the difference (gap) between current and previous mean similarity score. The bigger gap is used as threshold to remove images from the subject.}
    \label{fig:mislabeled_overview}
\end{figure*}
\begin{figure}[t]
    \centering
    \begin{subfigure}[b]{.3\columnwidth}
        \centering
        \includegraphics[width=\linewidth]{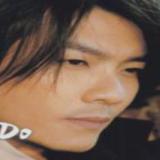}
    \end{subfigure} 
    \begin{subfigure}[b]{.3\columnwidth}
        \centering
        \includegraphics[width=\linewidth]{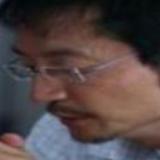}
    \end{subfigure}
    \caption{Example of images with large head pose that were removed in the pose cleaning process.}
    \label{fig:pose_cleaning}
\end{figure}

We believe that all of the problems mentioned in this section are common in any web-scraped dataset.
Thus, it is highly desirable that a cleaning procedure can be conducted on the initial web-scrapred dataset before it is used for training or testing models.

\section{Method}
\label{sec:method}
In this section we describe how we curated the AFD dataset to fix the problems described above.
The procedure used to curate the AFD dataset has also been carried out on the VGGFace2 test dataset, and we believe that it can be done with any other web-scraped dataset.

\subsection{Pose Cleaning}
As mentioned in the dataset analysis, to have a more comparable dataset, we perform pose filtering by deleting the images with large pose angle.
For this end, we use a commercial face API to predict pose angles.
We removed all images that have roll, yaw or pitch values more than $\left| 15\right|$ degrees.
Figure \ref{fig:pose_cleaning} shows two images that were removed in this pose cleaning process. 
After pose filtering, we also removed subjects that had fewer than 10 images as a result of the pose filtering. 
After this first step, the dataset contains 109,939 images of 1,960 subjects.

\subsection{Mislabelled Image Cleaning}
The second step in our cleaning approach is to remove images in each subject identity that do not belong to that subject.
We used ArcFace \cite{arcface} to extract face features of all images in the dataset resulting from the pose cleaning in the previous section.

Using the features extracted with ArcFace, for each image in a particular subject's folder, we calculated the mean value of the matching scores to the rest of the subjects' images.
If an image does not belong to the subject, its mean similarity with other images in that subject's folder will be much lower than the correctly labelled images. 
To separate the mislabelled images, we listed the mean similarity score of each image from highest to the lowest. 
Then, we calculated the difference between each ordered image mean similarity score (gap).
Using the biggest gap as a candidate threshold, all images that have a mean similarity lower than the image after the biggest gap are removed, as they are the mislabeled images.
Moreover, some images with bad quality will also be be excluded using this approach, since they had lower similarity scores with other images than the images with good quality. 
This procedure is repeated for each subject, where the biggest gap is calculated on a per-subject basis.
Figure \ref{fig:mislabeled_overview} summarizes this step.

\begin{figure}[t]
    \begin{subfigure}[b]{0.49\columnwidth}
        \centering
        \begin{subfigure}[b]{0.47\columnwidth}
            \centering
            \includegraphics[width=\linewidth]{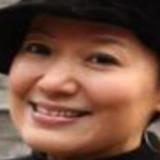}
        \end{subfigure}
        \begin{subfigure}[b]{0.47\columnwidth}
            \centering
            \includegraphics[width=\linewidth]{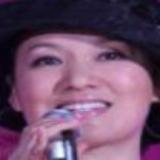}
        \end{subfigure}
        \caption{Images identified as \\``zhouyanhong''}
    \end{subfigure}
    \hfill
    \begin{subfigure}[b]{0.49\columnwidth}
        \centering
        \begin{subfigure}[b]{0.47\columnwidth}
            \centering
            \includegraphics[width=\linewidth]{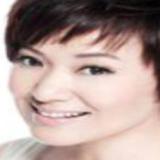}
        \end{subfigure}
        \begin{subfigure}[b]{0.47\columnwidth}
            \centering
            \includegraphics[width=\linewidth]{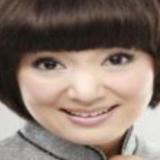}
        \end{subfigure}
        \caption{Images identified as \\``zhouyanhong1''}
    \end{subfigure}
    \caption{Example of two subjects that were merged in the merging step because they are the same person.}
    \label{fig:subject_merge}
\end{figure}

We subsequently deleted all subjects that had only one image, as they will not contribute to authentic distribution.
Finally, after the mislabeled image cleaning, and singleton removal, we ended with 101,293 images of 1927 subjects.

\subsection{Subject Merging}

After deleting the mislabelled images for each subject, each identity folder should contain images of only one person. However, we may still have instances of a person who has more than one identity folder.
To merge the subjects, we used a list of average similarity between identity folders.
To this end, we randomly chose 5 images from each identity folder to represent that identity.
If the subject had fewer than 5 images, we used all of their images.
Similar to last step, we used the features extracted by ArcFace to represent each image selected.
For each subject, we matched their 5 images against each other subjects, and then we calculated the mean of each subject against each other subject, creating a mean similarity of each subjects identity towards the others.

\begin{figure*}[t]
    \begin{subfigure}[b]{.12\textwidth}
        \centering
        \includegraphics[width=\linewidth]{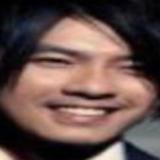}
        \caption{100\%}
    \end{subfigure}
    \begin{subfigure}[b]{.12\textwidth}
        \centering
        \includegraphics[width=\linewidth]{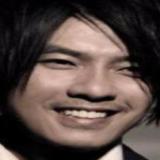}
        \caption{18.66\%}
    \end{subfigure}
    \begin{subfigure}[b]{.12\textwidth}
        \centering
        \includegraphics[width=\linewidth]{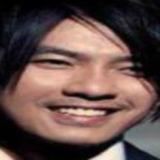}
        \caption{30.71\%}
    \end{subfigure}
    \begin{subfigure}[b]{.12\textwidth}
        \centering
        \includegraphics[width=\linewidth]{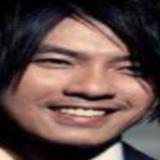}
        \caption{28.79\%}
    \end{subfigure}
    \begin{subfigure}[b]{.12\textwidth}
        \centering
        \includegraphics[width=\linewidth]{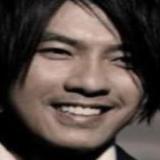}
        \caption{24.12\%}
    \end{subfigure}
    \begin{subfigure}[b]{.12\textwidth}
        \centering
        \includegraphics[width=\linewidth]{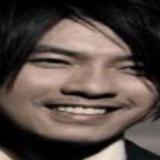}
        \caption{9.34\%}
    \end{subfigure}
    \begin{subfigure}[b]{.12\textwidth}
        \centering
        \includegraphics[width=\linewidth]{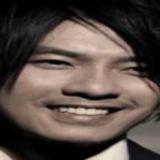}
        \caption{12.72\%}
    \end{subfigure}
    \begin{subfigure}[b]{.12\textwidth}
        \centering
        \includegraphics[width=\linewidth]{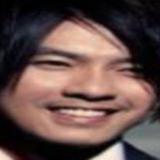}
        \caption{33.07\%}
    \end{subfigure}
    \caption{Example of near-duplicate images of subject ``adu''. Some of these images may seem identical, but they are not. The caption below each image represent the percentage of same pixels between each image and image (a).}
    \label{fig:near_duplicates}
\end{figure*}
\begin{figure*}[t]
    \centering
    \begin{subfigure}[b]{\textwidth}
        \centering
        \begin{subfigure}[b]{.24\textwidth}
            \centering
            \includegraphics[width=\linewidth]{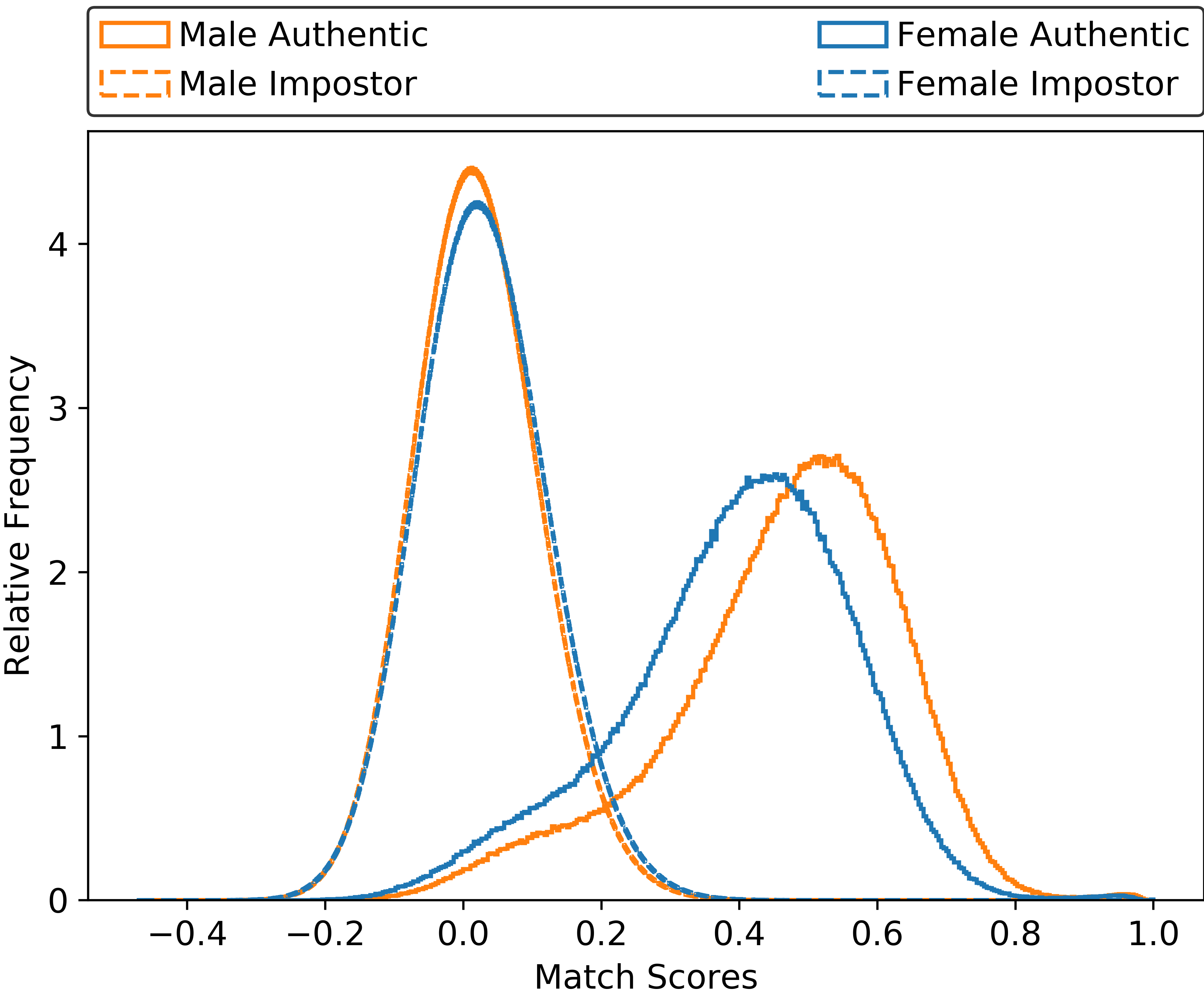}
        \end{subfigure}
        \hfill
        \begin{subfigure}[b]{.24\textwidth}
            \centering
            \includegraphics[width=\linewidth]{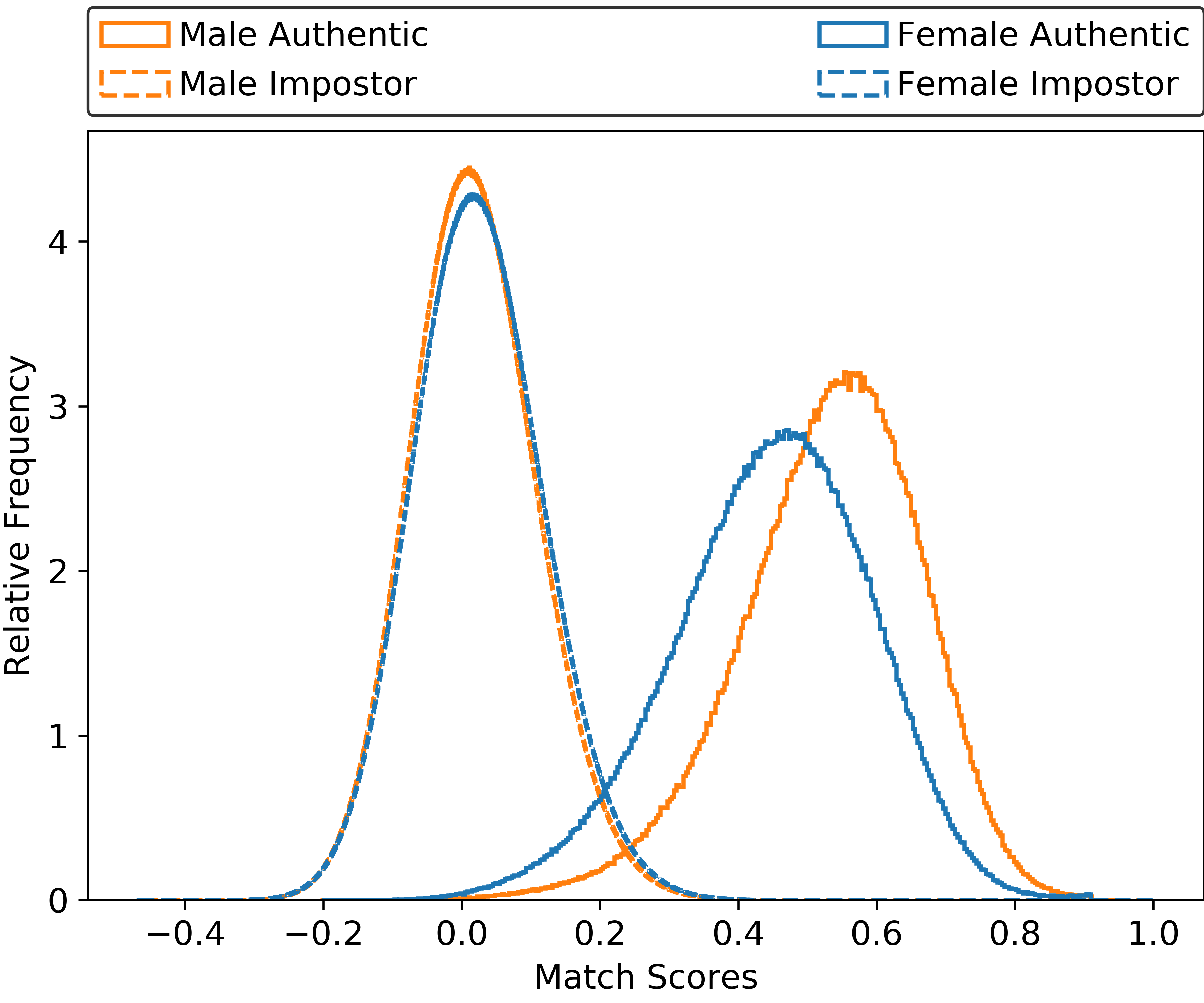}
        \end{subfigure}
        \hfill
        \begin{subfigure}[b]{.24\textwidth}
            \centering
            \includegraphics[width=\linewidth]{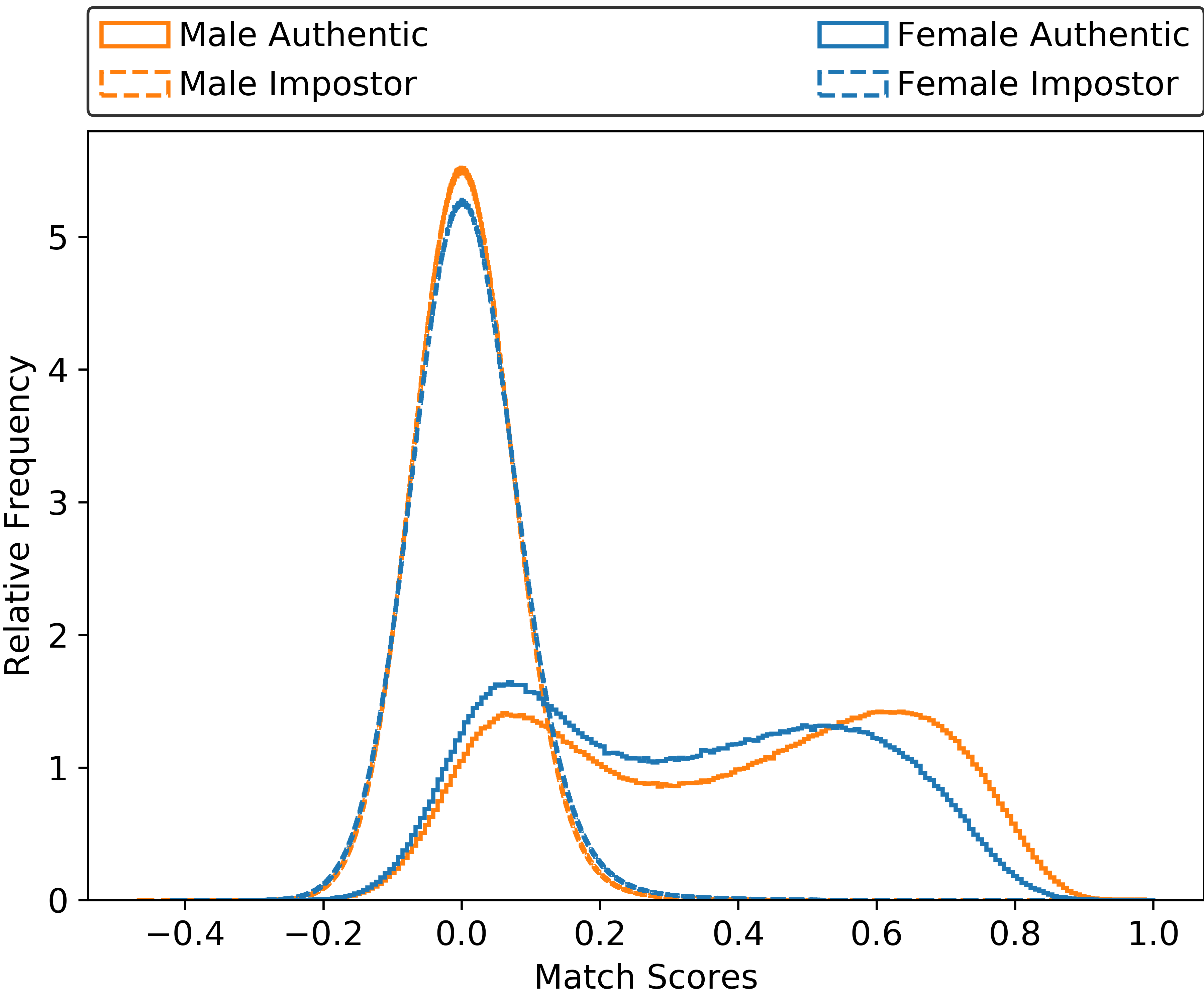}
        \end{subfigure}
        \hfill
        \begin{subfigure}[b]{.24\textwidth}
            \centering
            \includegraphics[width=\linewidth]{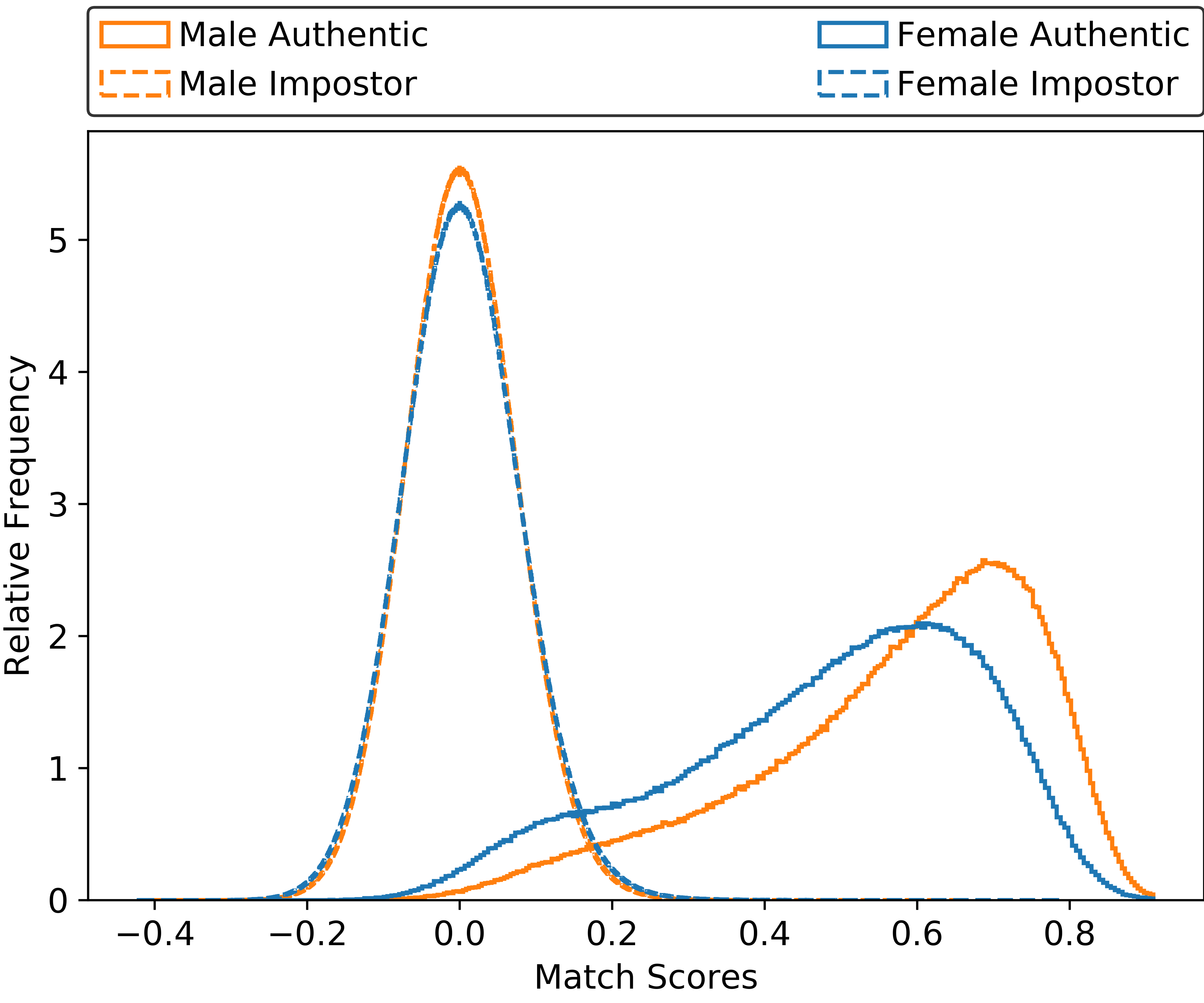}
        \end{subfigure}
    \end{subfigure}
    \begin{subfigure}[b]{\textwidth}
        \centering
        \begin{subfigure}[b]{.24\textwidth}
            \centering
            \includegraphics[width=\linewidth]{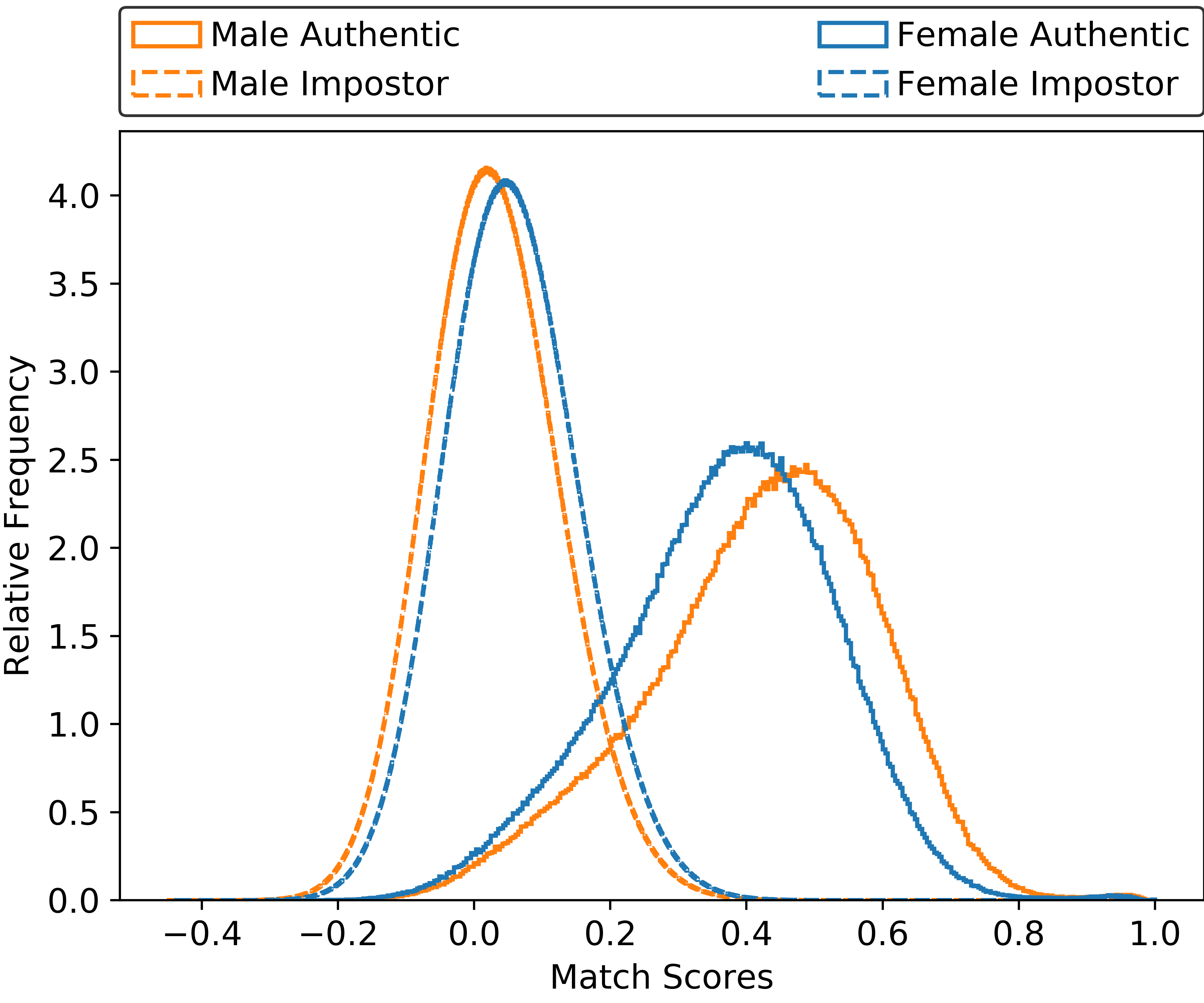}
            \caption{Original AFD}
        \end{subfigure}
        \hfill
        \begin{subfigure}[b]{.24\textwidth}
            \centering
            \includegraphics[width=\linewidth]{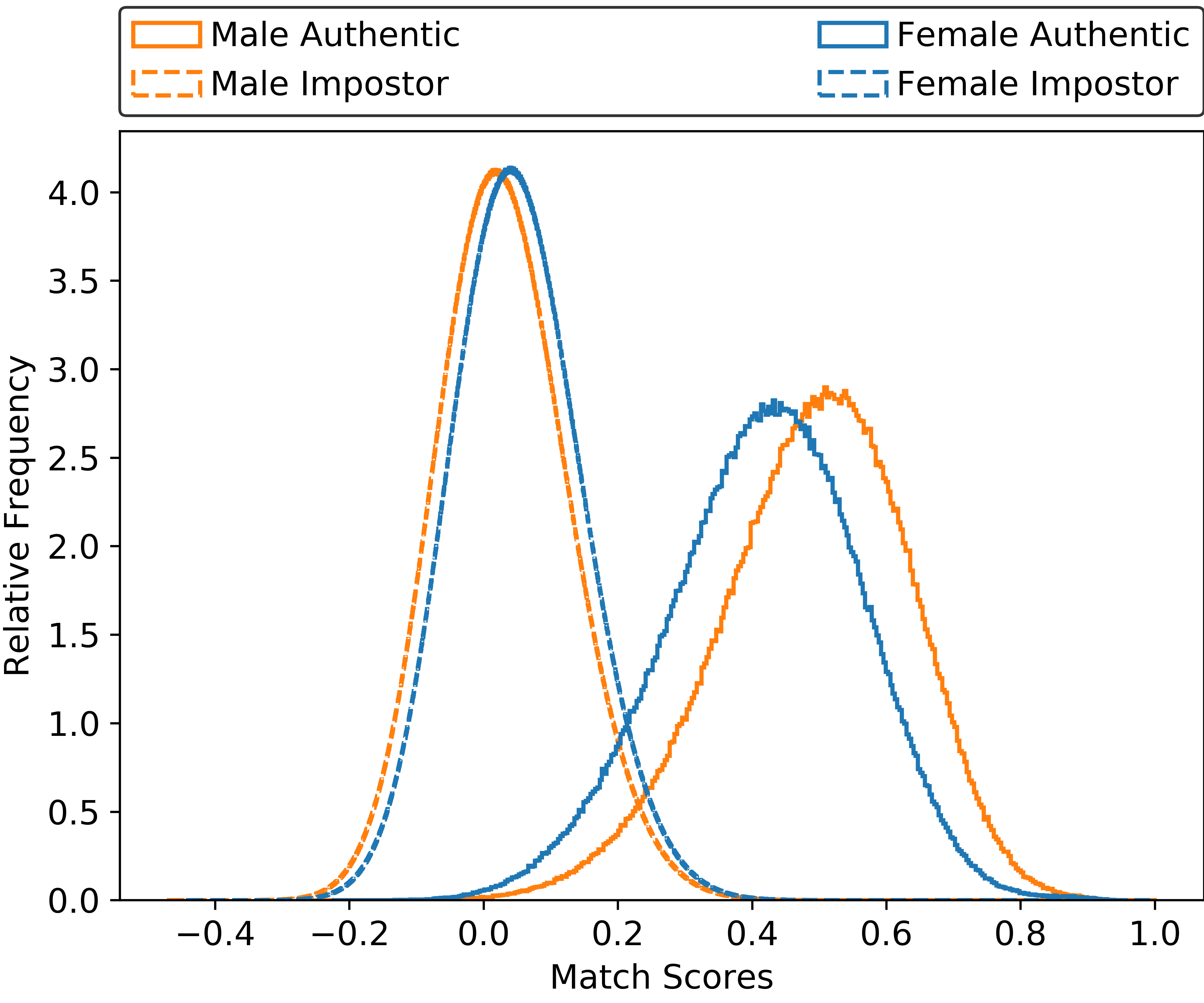}
            \caption{Curated AFD}
        \end{subfigure}
        \hfill
        \begin{subfigure}[b]{.24\textwidth}
            \centering
            \includegraphics[width=\linewidth]{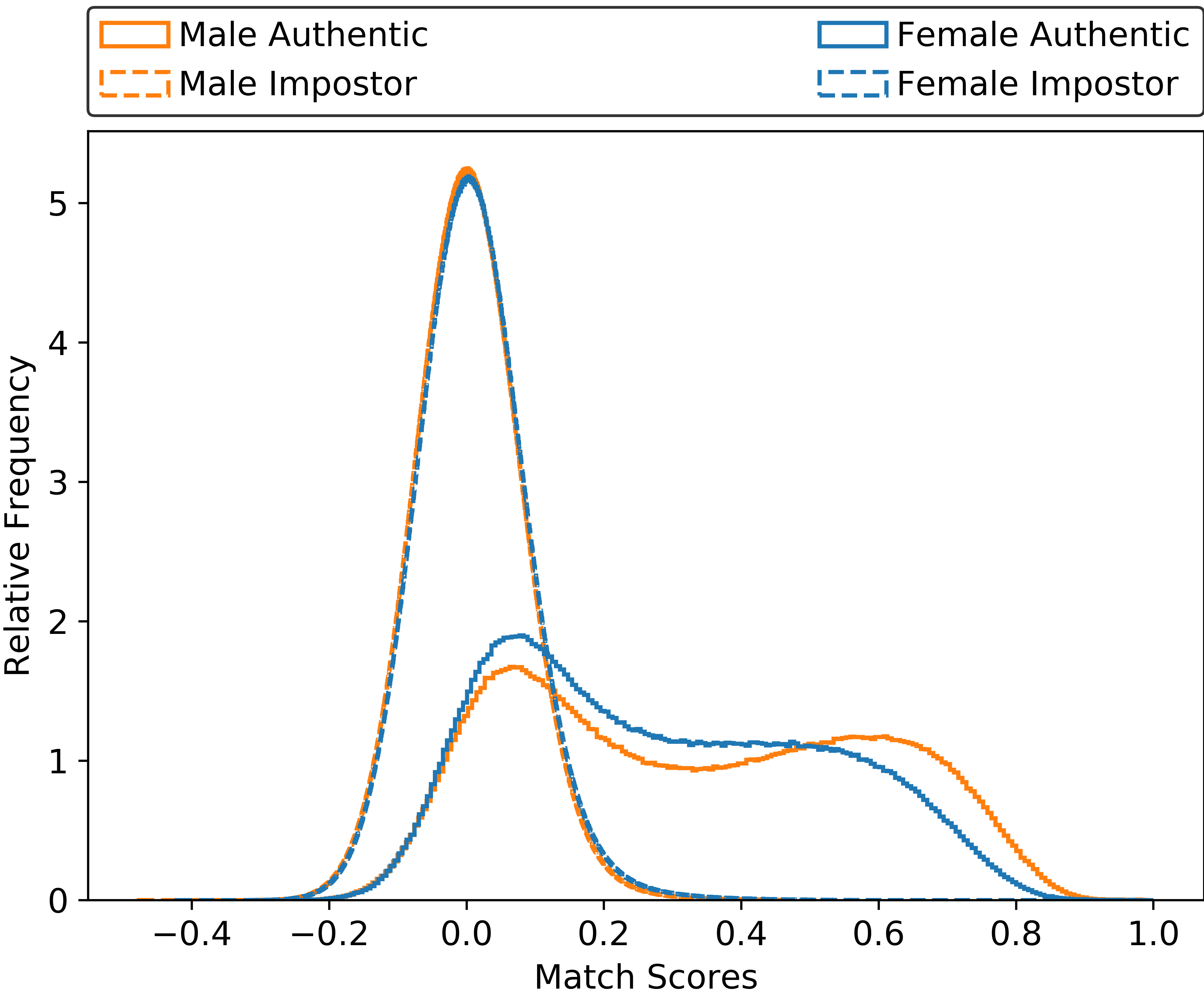}
            \caption{Original VGGFace2 Test}
        \end{subfigure}
        \hfill
        \begin{subfigure}[b]{.24\textwidth}
            \centering
            \includegraphics[width=\linewidth]{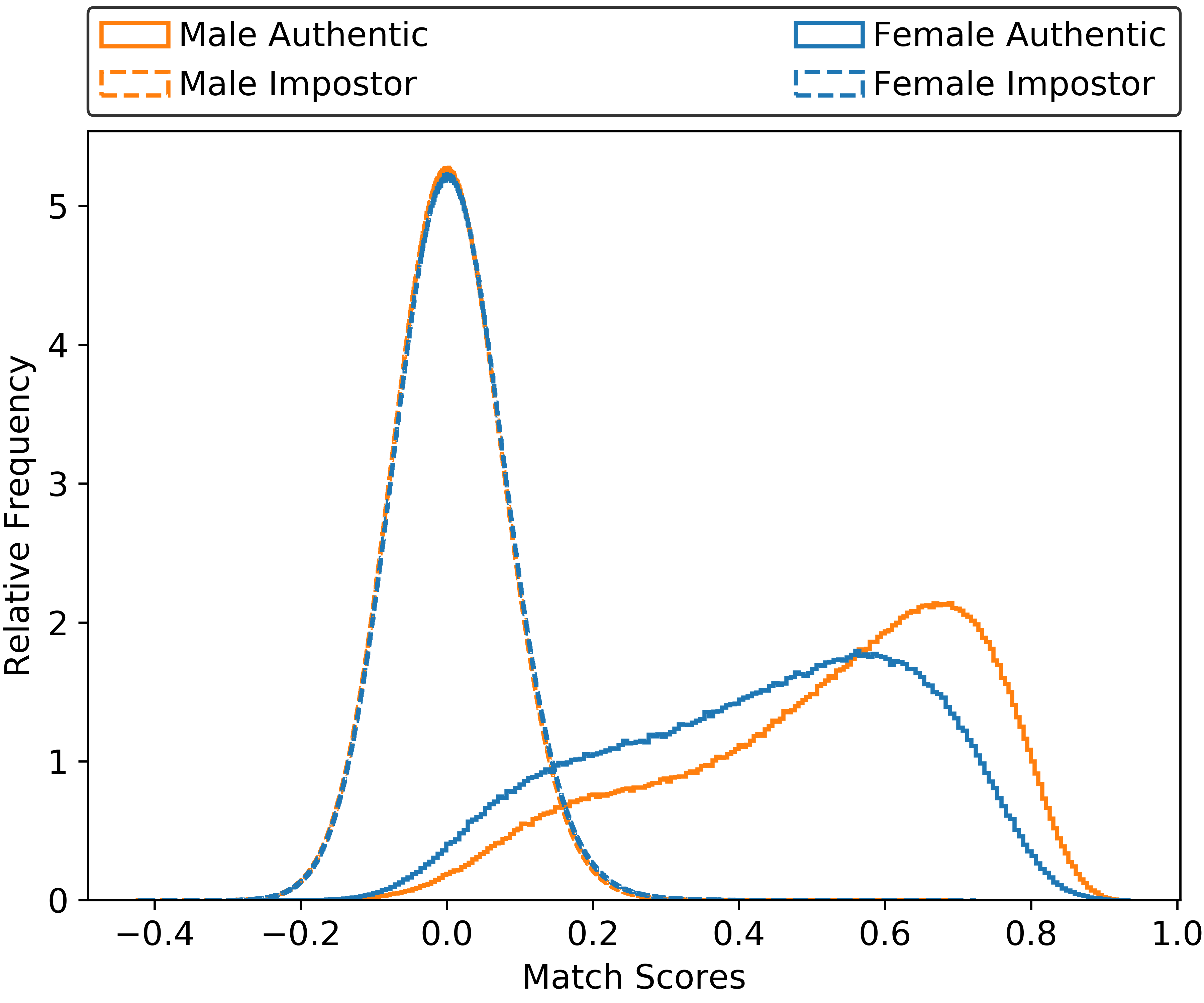}
            \caption{Curated VGGFace2 Test}
        \end{subfigure}
    \end{subfigure}
    \caption{Authentic and impostor distributions before and after cleaning. ArcFace shown on top, own matcher shown on bottom.}
    \label{fig:auth_imp}
\end{figure*}

Using the mean scores across subjects, we sort them from highest to lowest.
The pair of subject folders which had the highest similarity score was “shenxue” and “zhaohongbo”, which is the example shown in Section \ref{sec:data_analysis}. 
To decide whether a pair of subject folders with high similarity score are the same subject, we manually checked the subject images.
After manually inspection, we decided that any subject with a mean score lower than 0.25 was not the same subject.
For the 48 pairs of subjects with a score higher than 0.25, we merged all subjects that were the same, which were verified manually. 
Figure \ref{fig:subject_merge} shows a pair of subjects folders that were merged, ``zhouyanhong'' and ``zhouyanhong1''.
After merging subjects, there were 101,289 images and 1,917 subjects in total.

\subsection{Near-Duplicate Images Cleaning}
The final step in the proposed cleaning method is the near-duplicate image removal.
For each subject, we cross-matched all their images. We manually analyzed images for several subjects, and decided to use 0.91 as a threshold.
That is to say, we first chose a pivot image.
Then we matched the similarity between the pivot image and all other images. 
If the image has 0.91 or higher similarity score with the pivot image, we delete this image, as it is a near-duplicate.
After comparing this pivot image with all other images, we move to the next image, and repeat the process.
After all images left in the folder were chosen as pivot, all of the near-duplicates in that folder should be deleted. 
Figure \ref{fig:near_duplicates} show near-duplicates of a subject.
We deleted 9,642 images in this step. 
After deleting the subjects with fewer than 10 images, the final AFD dataset ended with 91,455 images of 1,878 subjects, which corresponds to 911 males with 42,134 images, and 967 females with 49,321 images.

\begin{figure*}[t]
    \centering
    \begin{subfigure}[b]{\textwidth}
        \centering
        \begin{subfigure}[b]{.24\textwidth}
            \centering
            \includegraphics[width=\linewidth]{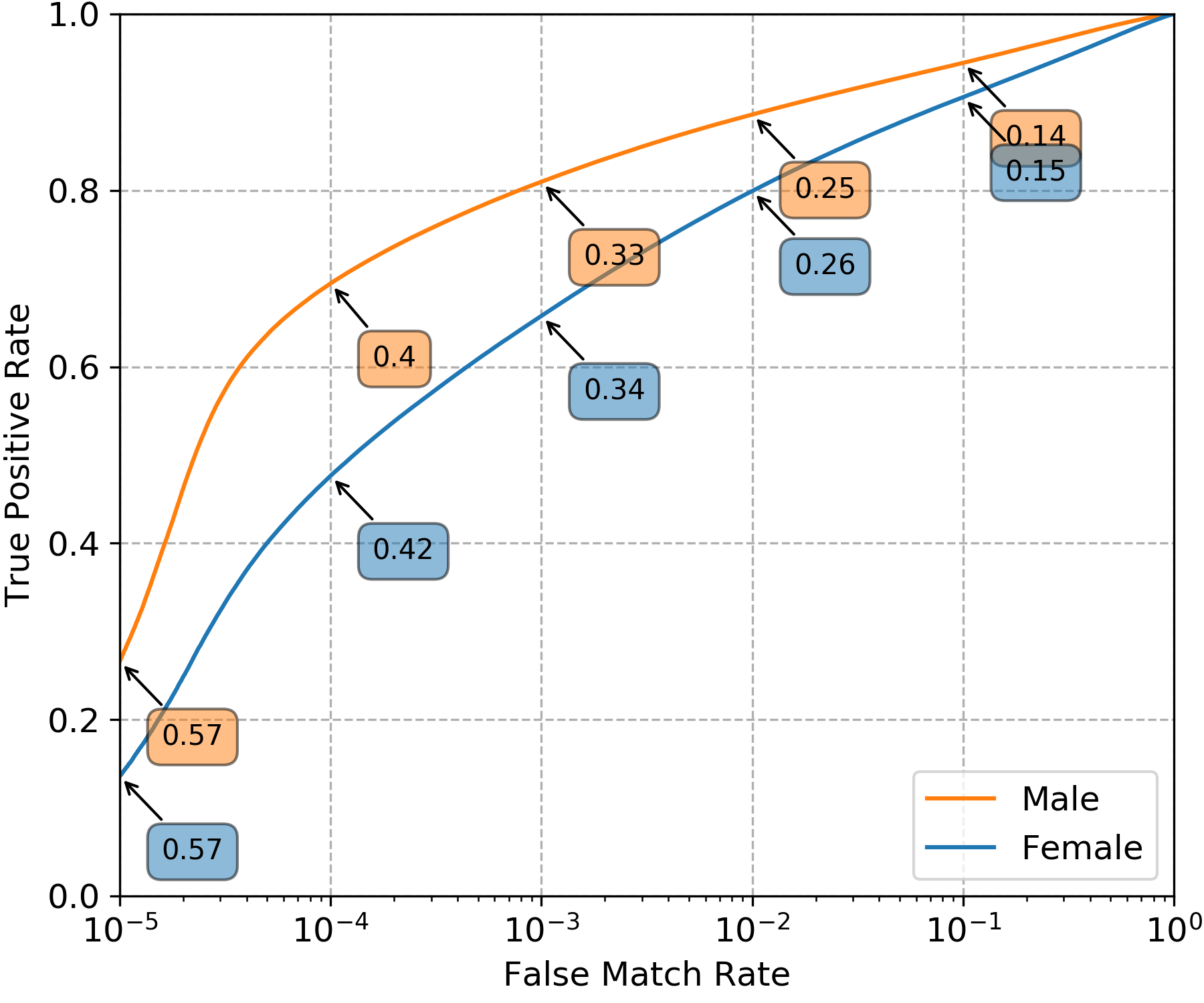}
        \end{subfigure}
        \hfill
        \begin{subfigure}[b]{.24\textwidth}
            \centering
            \includegraphics[width=\linewidth]{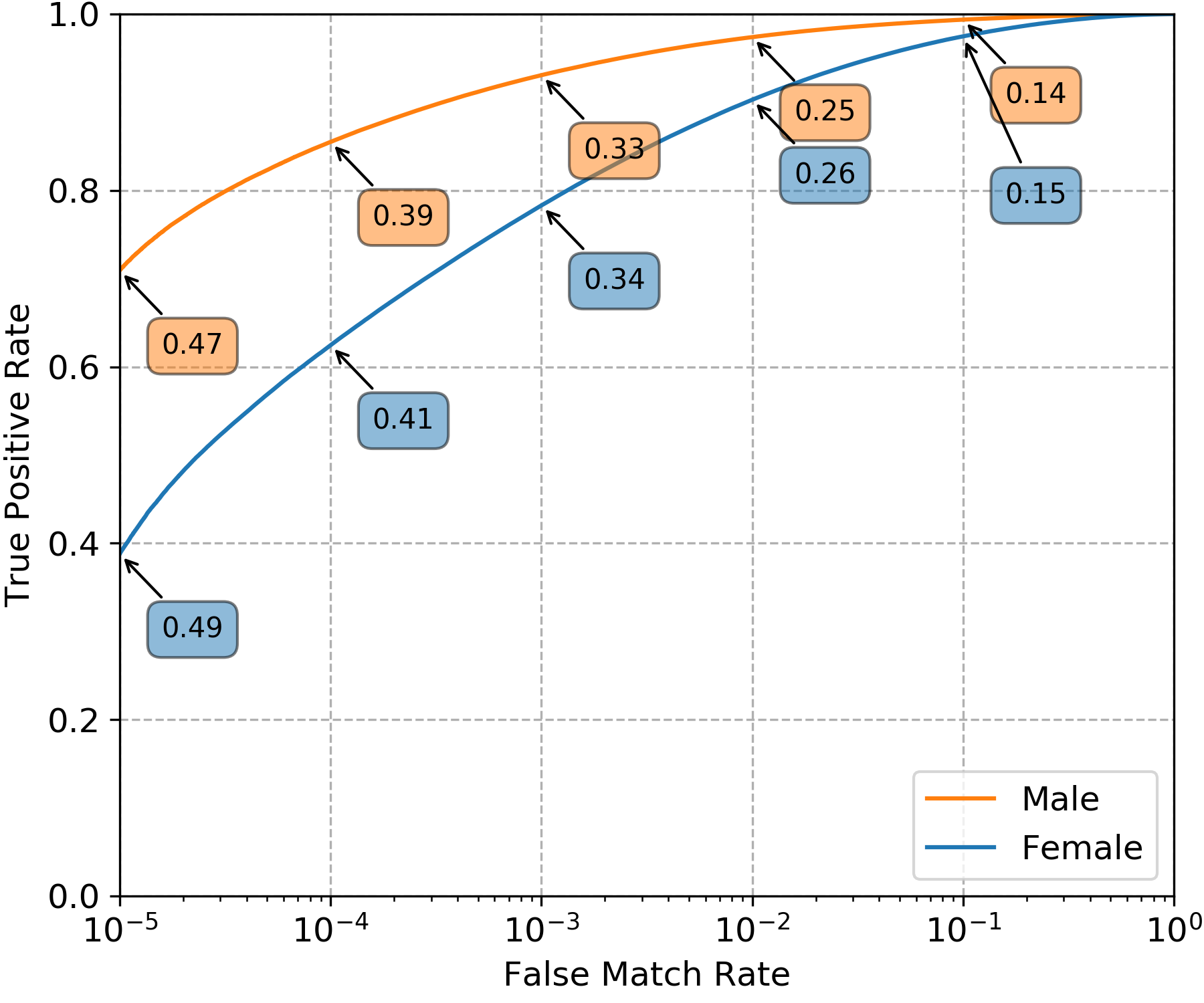}
        \end{subfigure}
        \hfill
        \begin{subfigure}[b]{.24\textwidth}
            \centering
            \includegraphics[width=\linewidth]{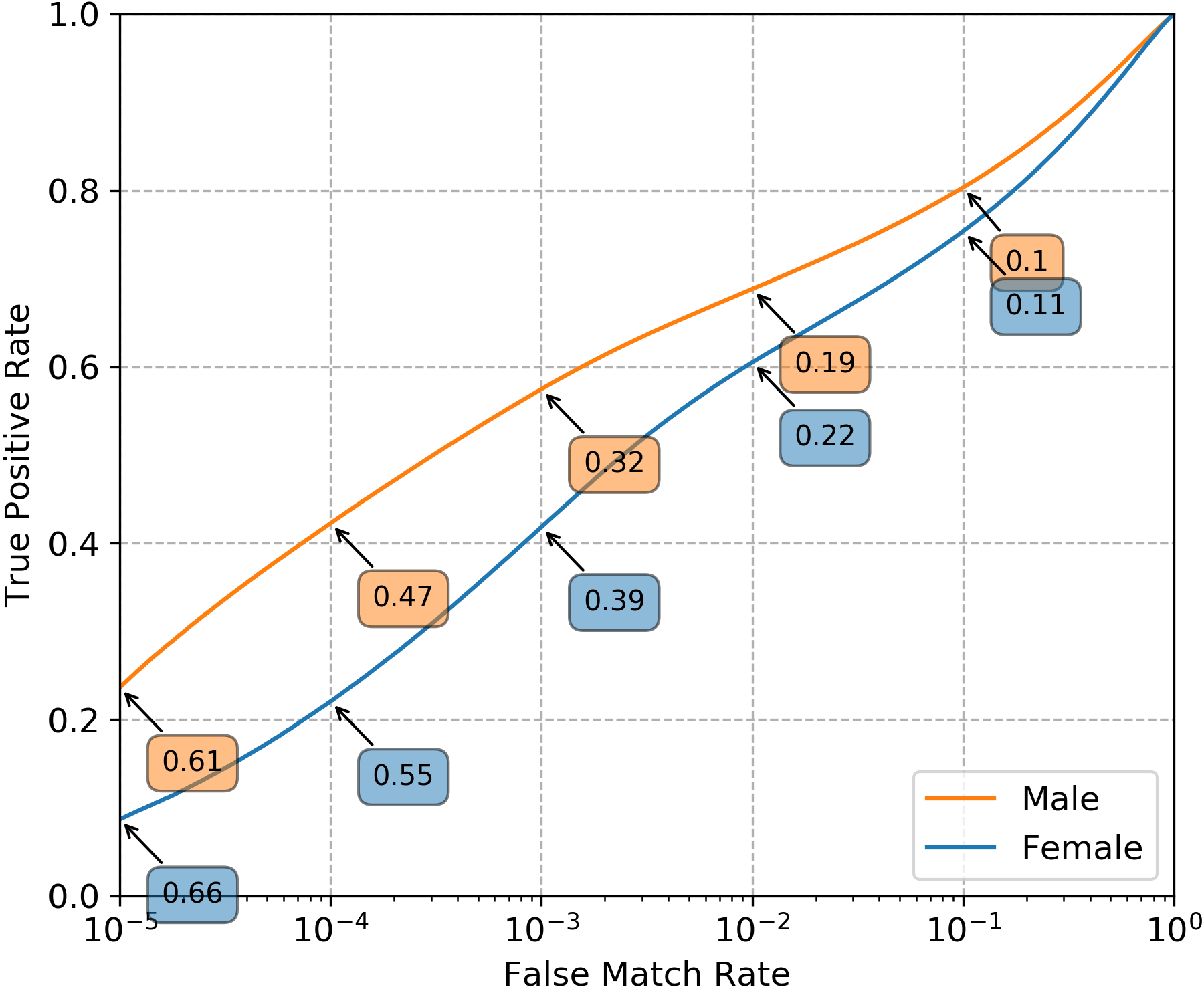}
        \end{subfigure}
        \hfill
        \begin{subfigure}[b]{.24\textwidth}
            \centering
            \includegraphics[width=\linewidth]{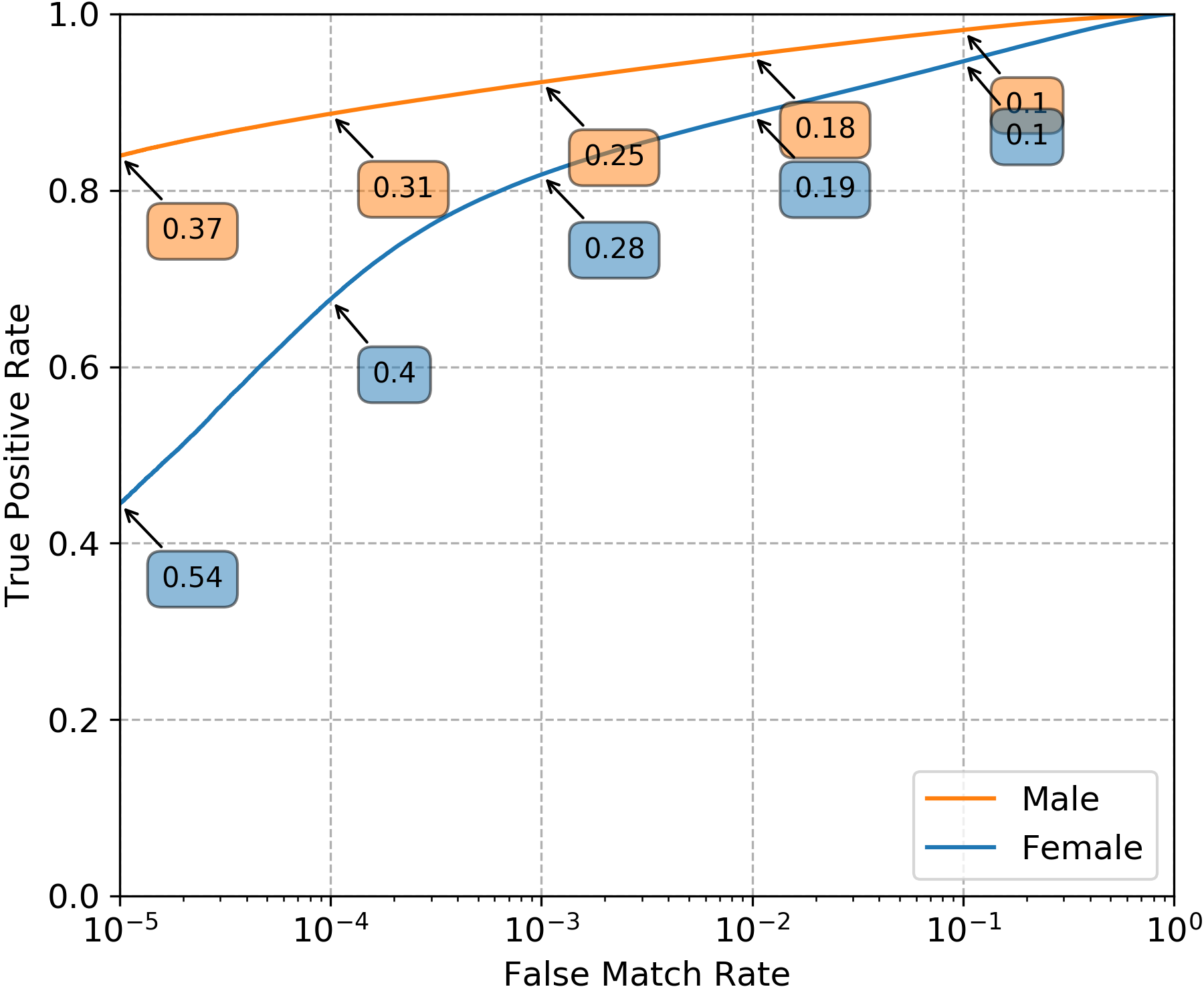}
        \end{subfigure}
    \end{subfigure}
    \begin{subfigure}[b]{\textwidth}
        \centering
        \begin{subfigure}[b]{.24\textwidth}
            \centering
            \includegraphics[width=\linewidth]{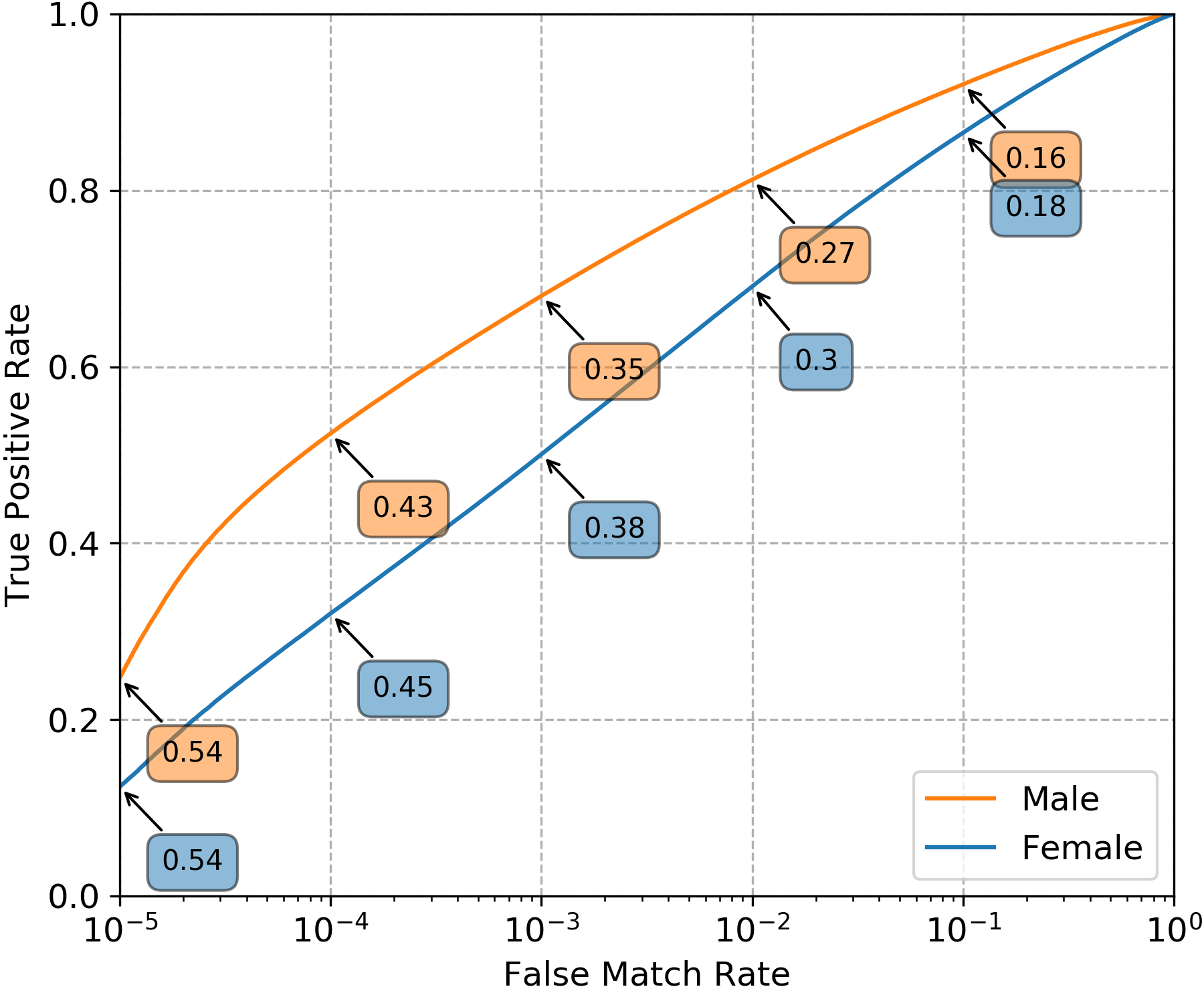}
            \caption{Original AFD}
        \end{subfigure}
        \hfill
        \begin{subfigure}[b]{.24\textwidth}
            \centering
            \includegraphics[width=\linewidth]{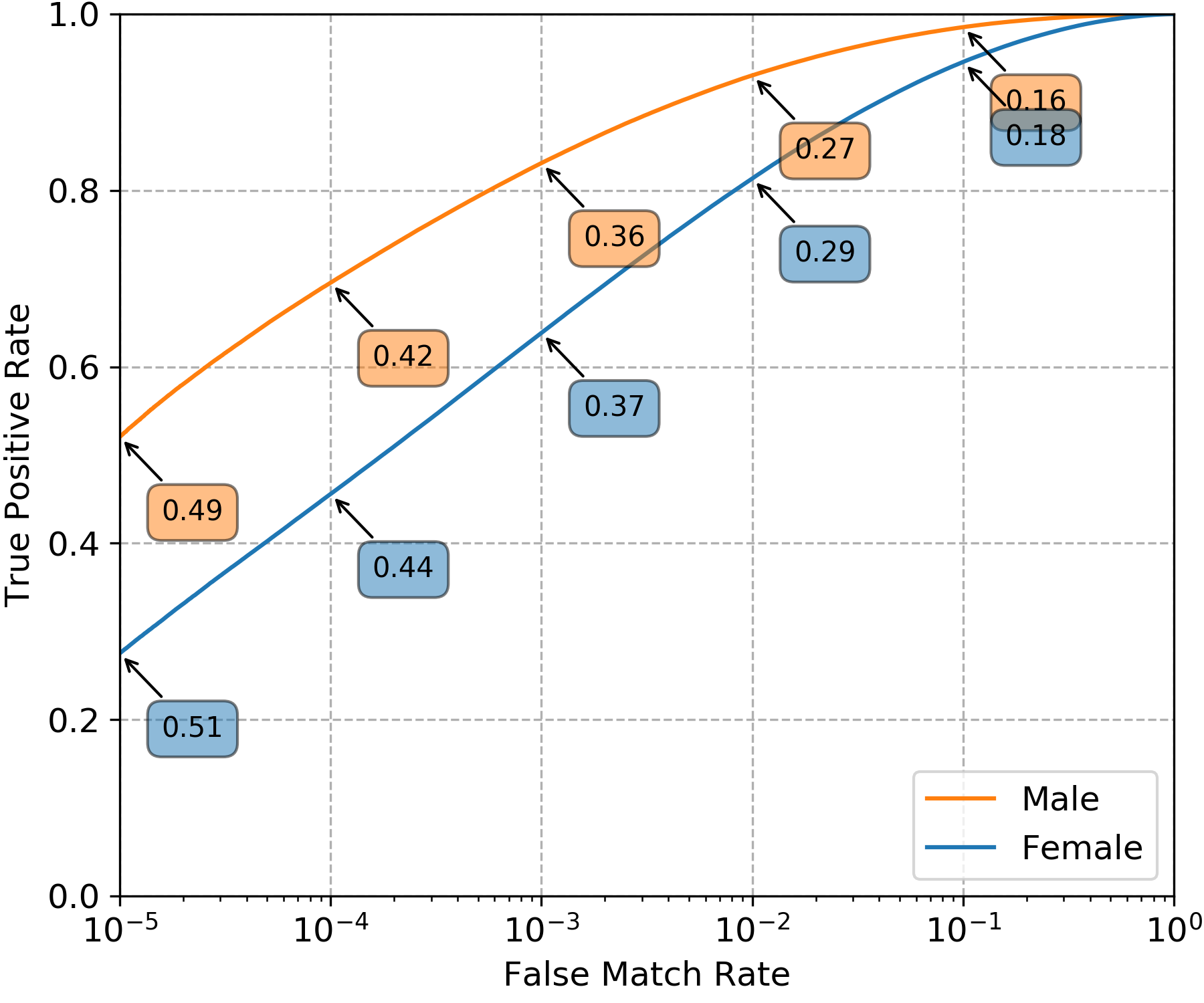}
            \caption{Curated AFD}
        \end{subfigure}
        \hfill
        \begin{subfigure}[b]{.24\textwidth}
            \centering
            \includegraphics[width=\linewidth]{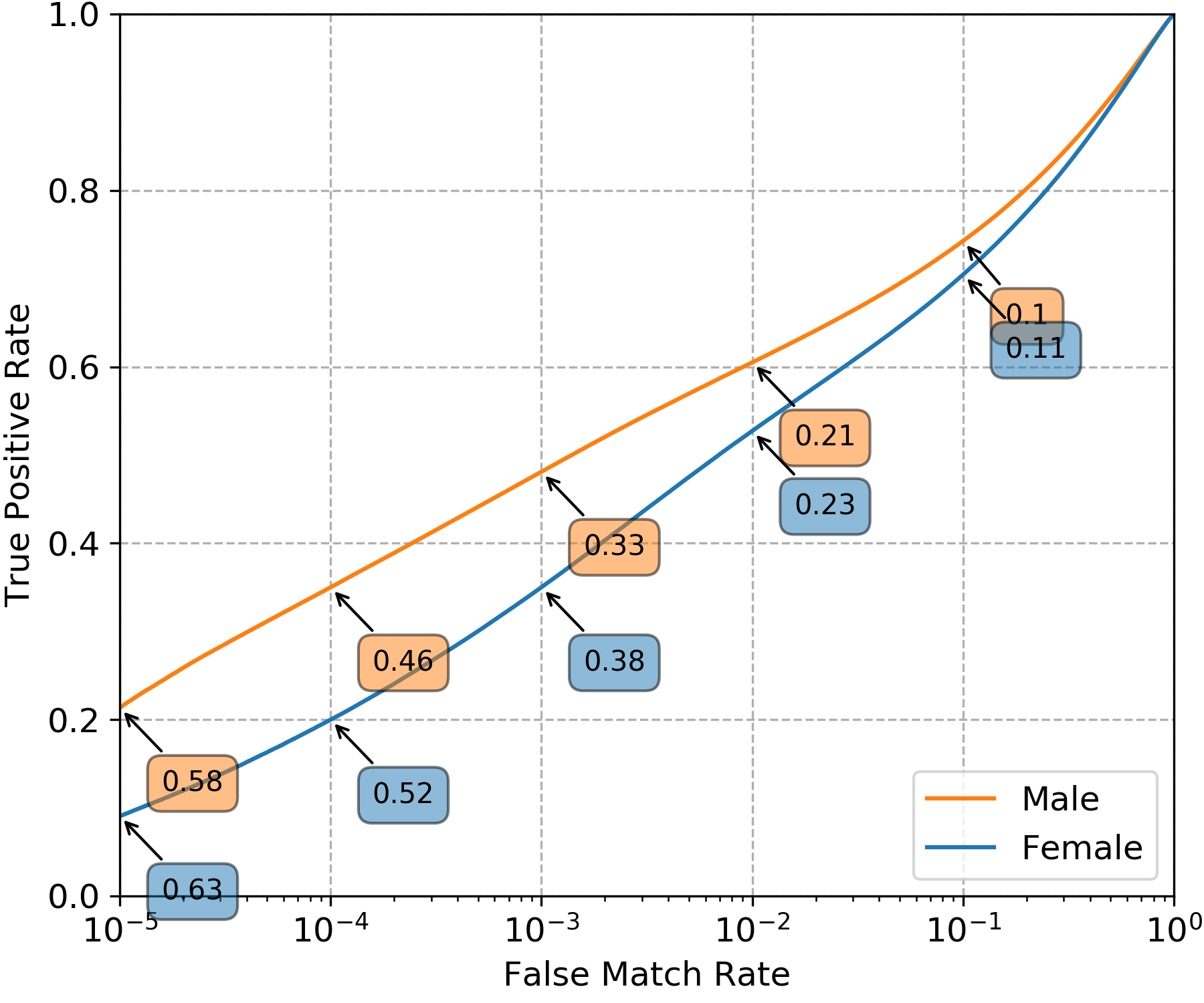}
            \caption{Original VGGFace2 Test}
        \end{subfigure}
        \hfill
        \begin{subfigure}[b]{.24\textwidth}
            \centering
            \includegraphics[width=\linewidth]{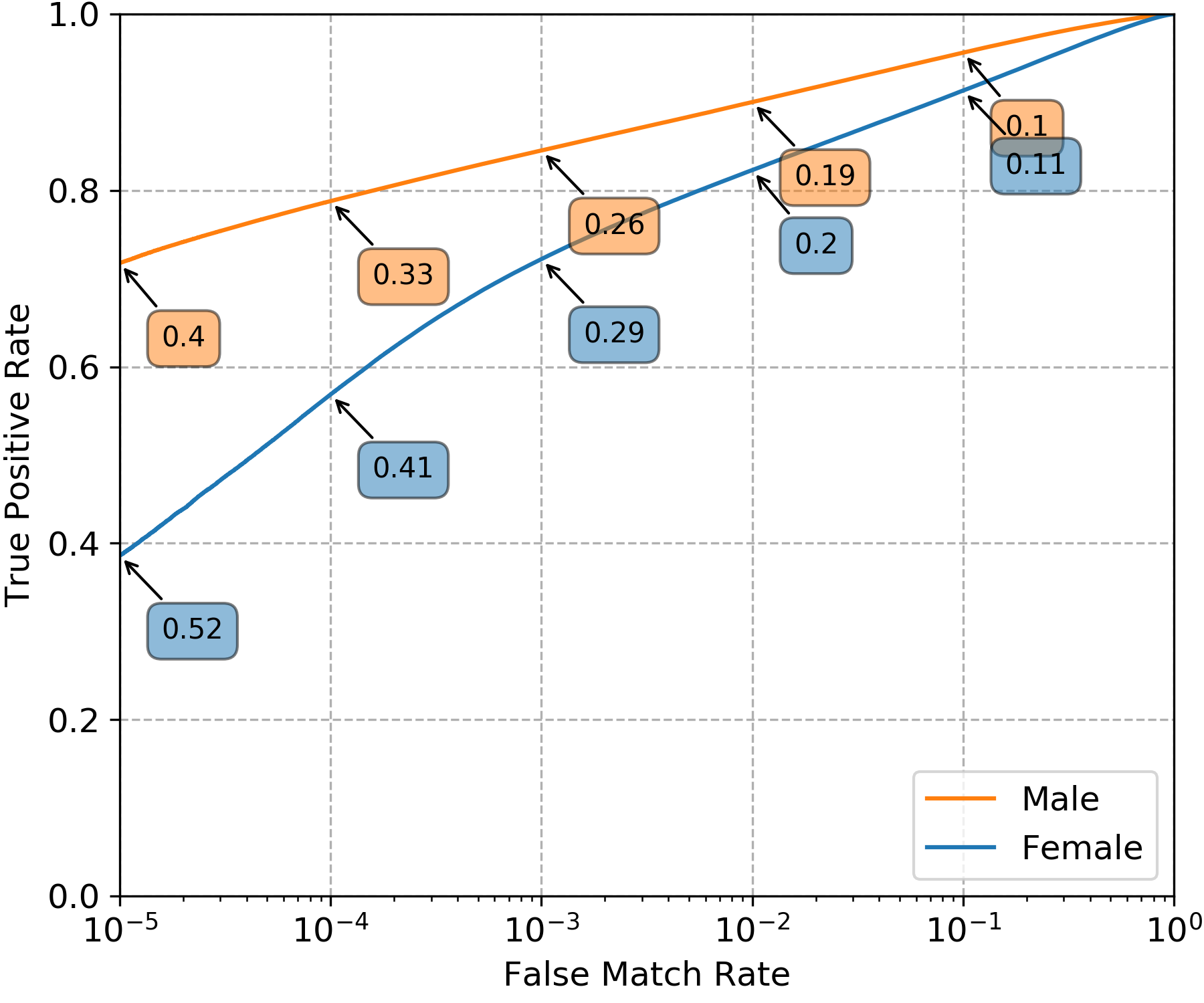}
            \caption{Curated VGGFace2 Test}
        \end{subfigure}
    \end{subfigure}
    \caption{ROC curves before and after cleaning. ArcFace shown on top, own matcher shown on bottom. Annotated values correspond to thresholds used to achieved a specific false match rate.}
    \label{fig:roc}
\end{figure*}

\section{Experiments}
In this section we investigate if the additional curation changes the result of face recognition  accuracy on the dataset.
However, by improving the accuracy, we mean getting what is supposed to be the real accuracy, since removing near- duplicates will lower the accuracy, but will make the accuracy estimate more realistic.

To validade the proposed method, along with ArcFace (that was the main tool for cleaning), we used a second matcher.
We trained the second matcher on a gender balanced subset of the MS1M-V2 dataset (which is a curated version of \cite{ms1_celeb}) using the ResNet-50 network and combined margin loss.
The combined margin loss, which was proposed by the authors of ArcFace \cite{arcface}, combines three losses: SphereFace \cite{sphereface}, CosFace \cite{cosface}, and ArcFace \cite{arcface}.
This matcher represents state-of-the-art performance, and for simplicity we call it ``own matcher''.

Along with adding a second matcher, that was not used during the cleaning procedure, we test the proposed cleaning method on a second dataset.
We repeat the steps on Section \ref{sec:method} on the VGGFace2 test dataset with the same thresholds\cite{vggface2}.
The initial VGGFace2 test dataset contains 101,752 images of 298 males, and 67,644 images of 202 females.
VGGFace2 has all the problems mentioned before, and as pointed out in~\cite{albiero2020does}, the test part of the dataset also has additional problems.
In particular, VGGFace2 test set has an age template matching procedure, where subject templates were created using 5 images of each subject to represent either young or old ages.
However, as reported in~\cite{albiero2020does}, in an extreme case, none of the images in a subject template were from the correct subject.
After curation, the VGGFace2 test dataset ended with 29,809 images of 295 males, and 21,469 images of 193 females.

As the original datasets contain more than 100,000 images for each gender, matching all against all is unfeasible, thus the results shown in this section for the original datasets were collected using randomly $1/3$ of the data available.

\subsection{Authentic and Impostor Distribution Comparison}

Figure \ref{fig:auth_imp} shows the authentic and impostor distribution for the AFD and VGGFace2 test datasets before and after the curation.
For the AFD dataset, we can observe that the high peak on the authentic distribution (near-duplicate images) is gone after the curation.
The authentic match scores closer to zero (mislabeled images) are reduced significantly.
On the other hand, the impostor distribution does not show big changes.

For the VGGFace2 test dataset, the authentic distribution which had a really strong peak close to match scores of zero (mislabeled images), moved to higher values.
We also observe that the authentic distribution on VGGFace2 test is multi-modal, which is not usual.
However, after cleaning, it become ``less multi-modal''.
As observed in the AFD dataset, the impostor distribution of VGGFace2 test does not show big changes, except that the impostor distribution is wider on both matchers.

In all cases, the female authentic distribution is worse than the male.
For the impostor distribution, the AFD dataset shows slightly worse distribution on the ArcFace matcher, and worse distributions on our own matcher.
For the VGGFace2 test dataset, the ArcFace matcher has a more similar impostor distribution between genders, and our own matcher shows almost the same distribution for males and females.

%
%

\subsection{ROC Comparison}
The receiver operating characteristic (ROC) curves for both datasets before and after curation are shown in Figure \ref{fig:roc}.
We observed that all the thresholds used to achieve a particular false match rate (FMR) drop significantly after the curation.
Also, both datasets show better ROC curves after the curation.
With a FMR of 1-in-100,000, on the AFD dataset using our own matcher, males true positive rate (TPR) went from 26.59\% to 70.92\%, and females went from 13.57\% to 38.78\%.
On the VGGFace2 test dataset, males TPR increased from 23.62\% to 83.93\%, and females increased from 8.69\% to 44.49\%.

Compared to males, females show higher thresholds on both datasets, before and after curation.
Also, before curation, males and females show a smaller difference in the ROC curve.
After curation, this difference is increased, with females having always lower performance.. 

\section{Conclusion}

We showed that web-scraped in-the-wild face datasets can have many types of problems, including mislabeled identities, flipped identity labels, low quality images, faces with high pose angles, multiple identities for the same person and near-duplicate images.
In this paper we presented a method to remove these problems so that the dataset can be used for testing the accuracy of face recognition and comparing results with more controlled datasets.
We explain the details of the approach using the AFD dataset as an example, but the proposed approach can be performed in any web-scraped face dataset, as we reproduce the method on a second dataset, VGGFace2.
Moreover, we will release both datasets for the research community.
In particular, AFD will be the largest available heavily-curated Asian face dataset.

Our experiments showed that the curation mostly affected the authentic distribution on both datasets, for both matchers.
The accuracy difference on the datasets before and after cleaning is big.
We showed that before cleaning, men have higher accuracy than females, and that after cleaning, this difference is increased significantly, even though both groups undergo the same cleaning steps.
The pattern of males having higher accuracy than females is mostly uniform across the literature \cite{frvt, Lui2009, Beveridge2009, frgc, Grother2010, Klare2012, Grother2017, cook2018, Lu2018, frvt3, albiero2020analysis, albiero2019how}.
%
%

\section*{Acknowledgment}
The authors would like to thank Kristen Haubold, Beth Marchant, and Erica Price, participants in the National Science Foundation ``Research Experiences for Teachers'' site (CNS 1855278) at the University of Notre Dame in the summer of 2019 for their early efforts on conceptualizing this project.

{
\bibliographystyle{ieee}
\bibliography{refs}
}
\end{document}